# Using the Random Sprays Retinex algorithm for global illumination estimation

Nikola Banić and Sven Lončarić
University of Zagreb, Faculty of Electrical Engineering and Computing
Unska 3, 10000 Zagreb, Croatia
E-mail: {nikola.banic, sven.loncaric}@fer.hr

*Abstract*—In this paper the use of Random Sprays Retinex (RSR) algorithm for global illumination estimation is proposed and its feasibility tested. Like other algorithms based on the Retinex model, RSR also provides local illumination estimation and brightness adjustment for each pixel and it is faster than other path-wise Retinex algorithms. As the assumption of the uniform illumination holds in many cases, it should be possible to use the mean of local illumination estimations of RSR as a global illumination estimation for images with (assumed) uniform illumination allowing also the accuracy to be easily measured. Therefore we propose a method for estimating global illumination estimation based on local RSR results. To our best knowledge this is the first time that RSR algorithm is used to obtain global illumination estimation. For our tests we use a publicly available color constancy image database for testing. The results are presented and discussed and it turns out that the proposed method outperforms many existing unsupervised color constancy algorithms. The source code is available at http://www.fer.unizg.hr/ipg/resources/color_constancy/.

## I. Introduction

A well known feature of human vision system (HVS) is its ability to recognize the color of objects under variable illumination when this color depends on the color of the light source [1]. An example of an object under different light sources can be seen on Fig. 1. This feature is called color constancy and achieving it computationally can significantly enhance the quality of digital images. Even though the HVS has generally no problem with it, computational color constancy is an ill-posed problem and assumptions have to be made for color constancy algorithms. Some of these assumptions include color distribution, uniform illumination, presence of white patches etc. After taking the Lambertian and one single light source assumption, the dependence of observed color of the light source **e** on the light source $I(\lambda)$ and camera sensitivity function $\rho(\lambda)$, which are both unknown, can be written as

$$\mathbf{e} = \begin{pmatrix} e_R \\ e_G \\ e_B \end{pmatrix} = \int_\omega I(\lambda)\rho_c(\lambda)d\lambda \qquad (1)$$

and it represents the illumination estimation.

Examples of color constancy algorithms include the gray-world algorithm [2], shades of gray [3], grey-edge [4], gamut mapping [5], using neural networks [6], using high-level visual information (HLVI) [7], probabilistic algorithms [8] and combination of existing methods [9]. The result of all mentioned algorithms is a single vector representing the global illumination estimation which is than used in chromatic adaptation to create an appearance of another desired illumination. Algorithms like those based on gamut mapping have the advantage of greater accuracy, but they need to be trained first. Some simpler and unsupervised algorithms like Gray-world or Gray-edge are of lesser accuracy, but are easy to implement and have a low computation cost.

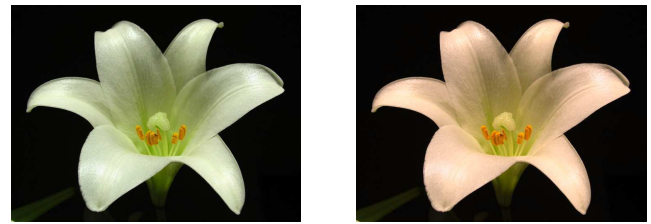

(a)                                (b)

Fig. 1: Same object under different illuminations.

A number of algorithms like [10] estimate the illumination locally and than combine multiple local results into one global thus also producing the global illumination estimation. In this paper we propose a similar method based on the Random Sprays Retinex (RSR) [11], an algorithm of Retinex model, which deals with *locality of color perception* [12], a phenomenon by which the HVS's perception of colors is influenced by the colors of adjacent areas in the scene. The algorithms of the Retinex model provide local white balancing and brightness adjustment producing so an enhanced image and not a single vector. If the assumption of uniform illumination is taken, then a single illumination estimation vector can be created from combined local estimations. RSR was chosen because it has the advantage of being faster than other path-wise Retinex algorithms [11].

This is the structure of the paper: in Section II a simple explanation of the Random Sprays Retinex algorithms is given, in Section III our proposed method for global illumination estimation is explained and in Section IV the evaluation results are presented and discussed.

## II. Random Sprays Retinex algorithm

The Random Sprays Retinex algorithm was developed by taking into consideration the mathematical description of

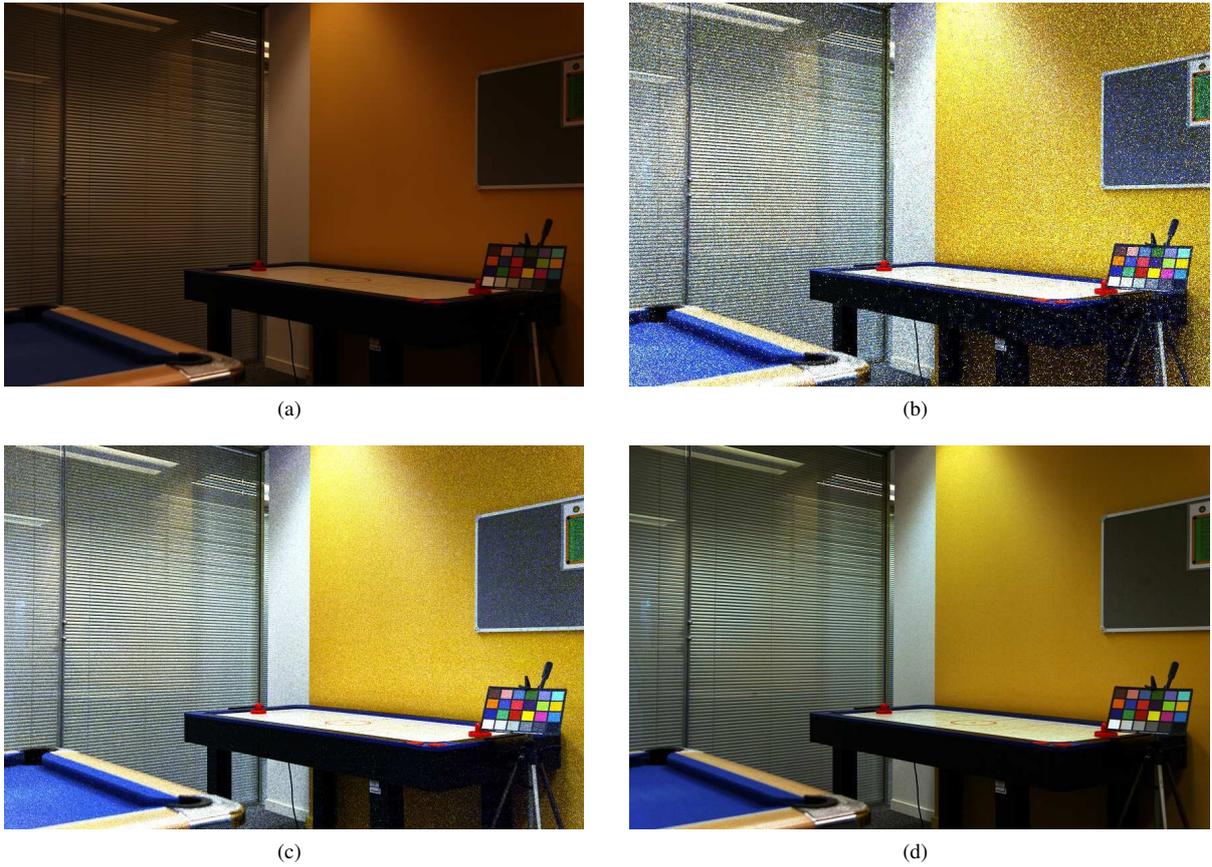

Fig. 2: Examples of images for various RSR parameters: (a) original image from the ColorChecker database, (b) $N=1, n=16$, (c) $N=5, n=20$, (d) $N=20, n=400$.

Retinex provided in [13]. After simplifying the initial model, it can be proved that the lightness value of pixel $i$ for a given channel can be calculated by using this formula

$$L(i) = \frac{1}{N} \sum_{k=1}^{N} \frac{I(i)}{I(x_{H_k})} \qquad (2)$$

where $I(i)$ is the initial intensity of pixel $i$, $N$ is the number of paths and $x_{H_k}$ is the index of the pixel with the highest intensity along the $k$th path.

The next step towards RSR in [11] is to notice three reasons for which paths should be replaced with something else: they are redundant, their ordering is completely unimportant and they have inadequate topological dimension. This leads to use of 2-D objects as representations of pixel neighbourhood, which is taken into consideration when calculating the new pixel intensity. Random sprays are finally chosen as these 2-D objects leading to several parameters that need to be tuned.

The value of the spray radius is set to be equal to the diagonal of the image. The identity function is taken as the radial density function. The minimal number of sprays ($N$) and the minimal number of pixels per spray ($n$) representing a trade-off between results quality and computation cost were determined to be 20 and 400 respectively [11]. Fig. 2(a) shows a test image from the ColorChecker image database [8]. RSR processed images for various parameters are shown on Fig. 2(b), Fig. 2(c) and Fig. 2(d).

### III. PROPOSED METHOD

#### A. Basic idea

From the original image and the RSR resulting image for a pixel $i$ it is possible to calculate its relative intensity change for a given channel $c$ using the equation

$$C_c(i) = \frac{I_c(i)}{R_c(i)} \qquad (3)$$

where $C_c(i)$ is the intensity change of pixel $i$ for channel $c$, $I_c(i)$ the original intensity and $R_c(i)$ the intensity obtained by RSR. Considering the way it is calculated, the vector $\mathbf{p}(i) = [C_r(i), C_g(i), C_b(i)]^\top$ composed of one pixel intensity change element for each channel can be interpreted as RSR local illumination estimation and since it is not necessarily normalized, it also represents the local brightness adjustment. That means that $\mathbf{p}(i)$ can also be written as $\mathbf{p}(i) = w(i)\hat{\mathbf{p}}(i)$ where $w(i) = \|\mathbf{p}(i)\|$ is the norm of $\mathbf{p}(i)$ and $\hat{\mathbf{p}}(i)$ is the unit vector with the same direction as $\mathbf{p}(i)$. The merged result of Eq. 3 for all channels calculated by using the RSR result from Fig. 2(b) is shown on Fig. 3(a).

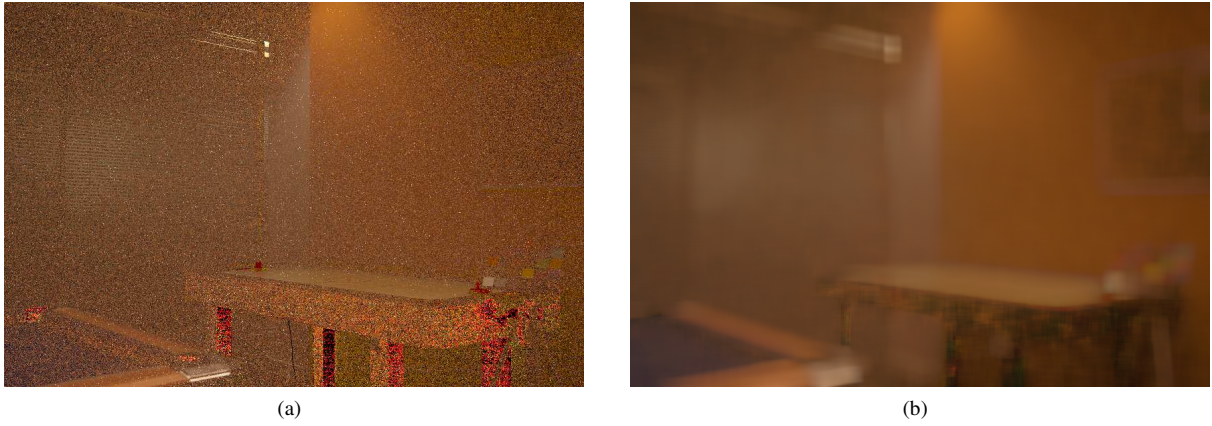

Fig. 3: (a) Local intensity changes for image shown on Fig. 2(a). (b) Local intensity changes after blurring the original and the RSR image.

As it is obvious that for some cases like the one shown on Fig. 3(a) there is a higher level of visible noise when visualising the intensity changes, it might be a good thing to try to lessen the noise in some way before further using the calculated changes. A good way of doing so is to apply a modification to Eq. 3:

$$C_{c,k}(i) = \frac{(I_c * k)(i)}{(R_c * k)(i)} \quad (4)$$

where $k$ is a chosen kernel and $*$ is the convolution operator. By applying Eq. 4 with an averaging kernel of size $25 \times 25$ instead of Eq. 3, the result shown on Fig. 3(a) turns into the one shown on Fig. 3(b).

A simple way of obtaining the global illumination estimation now is to calculate the vector

$$\mathbf{e} = \sum_{i=1}^{M} \mathbf{p}(i) = \sum_{i=1}^{M} w(i)\hat{\mathbf{p}}(i) \quad (5)$$

where $\mathbf{e}$ is the final global illumination estimation vector and $M$ is the number of pixels in the image. The division by $M$ is omitted since we are only interested in the direction of $\mathbf{e}$ and not its norm. It is obvious that vectors $\mathbf{p}(i)$ with greater corresponding value of $w(i)$ will have a greater impact on the final direction of vector $\mathbf{e}$ so $w(i)$ can also be interpreted as the weight. A simple alternative to Eq. 5 is to omit the weight $w(i)$:

$$\mathbf{e} = \sum_{i=1}^{M} \hat{\mathbf{p}}(i) \quad (6)$$

*B. Pixel sampling*

By looking at Fig. 3(b), it is clear that the difference between intensity changes of spatially close pixels is not large. This can be taken advantage of by not calculating $\mathbf{p}(i)$ for all pixels, but only for some of them. For that purpose the row step $r$ and the column step $c$ are introduced meaning that only every $c$th pixel only in every $r$th row is processed. By doing so, the computation cost is reduced, which can have an important impact on speed when greater values for $r$ and $c$ are used.

*C. Parameters*

The proposed method inherits all parameters from RSR, the two most important being the number of sprays $N$ and the size of individual sprays $n$. Additional parameters include filter kernel type and size and $r$ and $c$, the row step and column step respectively for choosing the pixels for which to estimate the illumination. As all of these parameters have a potential influence on the final direction, a tuning is necessary. Due to many possible combination of parameters, the parameters inherited from RSR and not mentioned in this subsection are set to values they were tuned to in [11]. Also in order to simplify further testing, only the averaging kernel was used due to its simplicity, low computation cost and no apparent disadvantage over other kernels after several simple tests (it even provided for a greater accuracy over the case when Gaussian kernels were used).

*D. Method's name*

As described in a previous subsection, the proposed method is designed to have an adjustable pixel sampling rate. As this allows the proposed method to "fly" over some pixels without visiting them, we decided to name the proposed method Color Sparrow (CS).

IV. EVALUATION AND RESULTS

*A. Image database and testing method*

For testing the proposed method and tuning the parameters the publicly available re-processed version [14] of the ColorChecker database described in [8] was used. It contains 568 different linear RGB images taken both outside and in closed space each of them containing a GretagMacbeth color checker used to calculate the groundtruth illumination data which is provided together with the images. The positions of the whole color checker and individual color patches of the color checker in each image are also provided, which allowes for the color checker to be simply masked out during the evaluation. As the error measure for white balancing the angle between RGB value of groundtruth illumination and the estimated illumination of an image was chosen. It should be noted that because the images in the Color Checker database are of medium variety [15], it might be necessary to retune the parameters by using some other, larger databases in order to obtain parameter values that would be a good choice for images outside the Color Checker database.

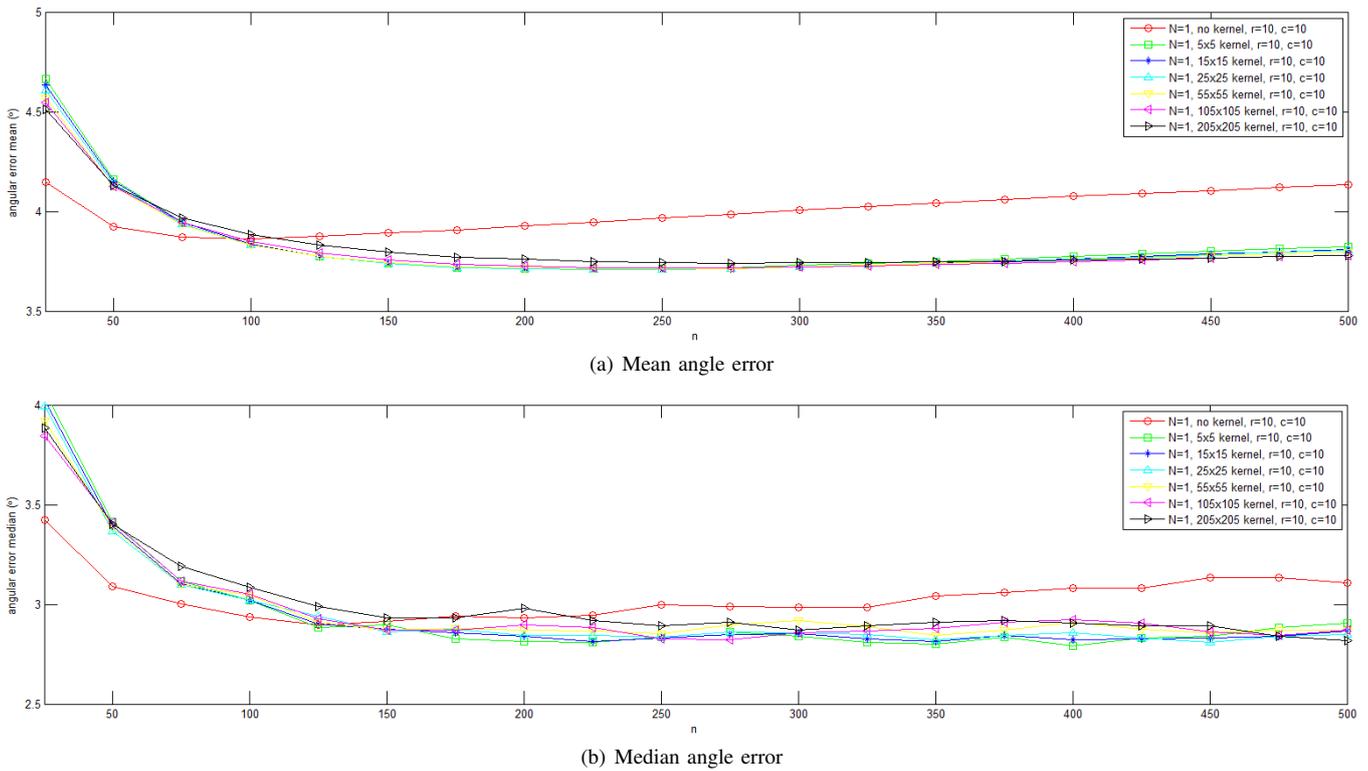

Fig. 4: Performance of several parameter settings with respect to different kernel sizes

Earlier in this paper two ways were proposed to estimate the direction of the global illumination and these are represented by Eq. 5 and Eq. 6. As in the performed experiments the former equation slightly outperformed the latter, all results mentioned in this paper were obtained by using Eq. 5, which means that there was no normalization of local intensity change vectors.

*B. Tuning the kernel size*

As explained before, the kernel type was fixed to averaging kernel. Fig. 4 shows two aspects of kernel size influence on method performance. All parameter settings have their $r$ and $c$ parameters set to 10 and $N$ was set to 1, while the value of $n$ varies. The reason to use only one value for $N$ is that raising it has insignificant impact on the performance. The graphs on Fig. 4 show a clear performance difference between application of Eq. 3, which uses no kernel, and Eq. 4. It is interesting to note that different kernel sizes have only slight impact on the mean angular error but this impact is greater on the median error. While the mean raises from $3.7°$ to $3.8°$ only when using the $205 \times 205$ kernel, the median raises from $2.8°$ to $2.9°$ when using kernels of size greater than $25 \times 25$. Therefore the kernel size should be in the range from $5 \times 5$ to $25 \times 25$ inclusive and in further tests the kernel size $5 \times 5$ is used.

*C. Tuning N and n parameters*

Both Fig. 5 and Fig. 4 show mean and median angle between groundtruth global illumination and global illumination estimation of CS calculated on Shi's images with various values of parameter $n$ for several fixed combinations of other parameters. It can be seen the lowest mean angular error of $3.7°$ and lowest median error of $2.8°$ are achieved for the value of $n$ between 200 and 250 and that this is almost invariant to values of other parameters. For that reason the result of parameter $n$ tuning is the value 225. As in previous tuning, in this case using greater values for parameter $N$ had also only slight impact and was therefore omitted. Performance for distinct image sizes was also tested by simply shrinking the original Shi's images and there was no significant difference in results.

*D. Tuning r and c parameters*

In order to retain the lowest achieved median and mean errors, the $r$ and $c$ parameter values can be raised up to 50 setting at the same time $n$ to 225 a shown on Fig. 5. It is interesting to mention that even setting $r$ and $c$ to 200 raises the mean angular error only to $3.8°$. However, as the median is less stable, the values of $r$ and $c$ should not exceed 50 in order to avoid a significant loss of accuracy as after this point the median angle error raises to $2.9°$.

*E. Comparison to other algorithms*

Table I shows performance of various methods. For the Standard Deviation Weighted Gray-world (SDWGW) the images were subdivided into 100 blocks. Color Sparrow was used with parameters $N = 1, n = 225, r = 50, c = 50$ and with an averaging kernel of size $5 \times 5$. The performance of other mentioned methods was taken from [16] where parameter values are also provided. The method described in [17] is not mentioned because the comparison would not be fair since this was not designed for images recorded under assumed single light source. It is possible to see that CS outperforms many methods and is therefore a suitable choice for white balancing.

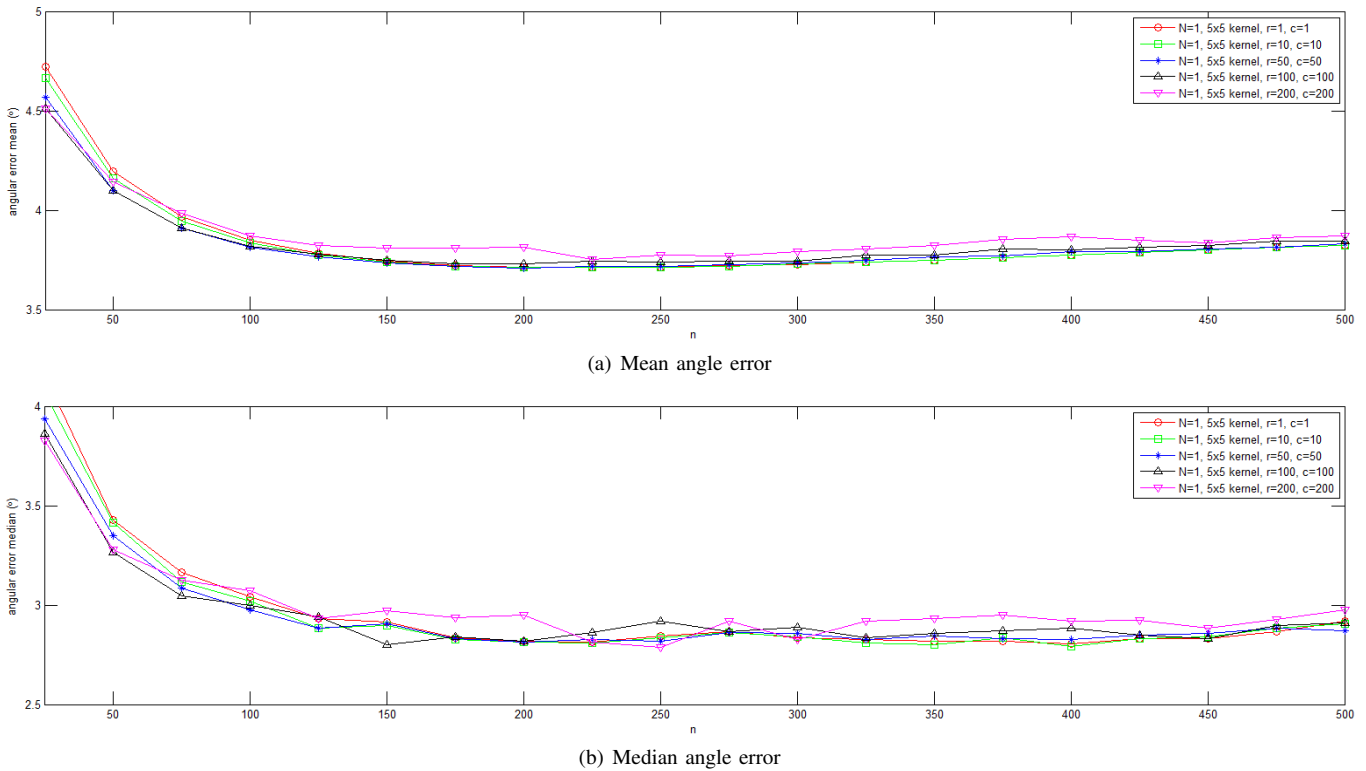

(a) Mean angle error

(b) Median angle error

Fig. 5: Performance of several parameter settings with respect to different sampling rates

TABLE I: Performance of different color constancy methods

| method | mean (°) | median (°) | trimean (°) | max (°) |
| --- | --- | --- | --- | --- |
| do nothing | 13.7 | 13.6 | 13.5 | 27.4 |
| Gray-world | 6.4 | 6.3 | 6.3 | 24.8 |
| SDWGW | 5.4 | 4.9 | 4.9 | 22.9 |
| Shades of gray | 4.9 | 4.0 | 4.2 | 22.4 |
| Gray-edge | 5.1 | 4.4 | 4.6 | 23.9 |
| Intersection-based Gamut | 4.2 | 2.3 | 2.9 | 24.2 |
| Pixel-based Gamut | 4.2 | 2.3 | 2.9 | 23.2 |
| HLVI | 3.5 | 2.5 | 2.6 | 25.2 |
| **proposed method** | 3.7 | 2.8 | 3.1 | 23.4 |

*F. Speed comparison*

As speed is an important factor property of white balancing algorithms, especially in digital cameras that have lower computation power, a speed test was also performed for CS. In order to compare the result of the speed test to something, a speed test was also performed for the Gray-world algorithm because it is one of the simplest and fastest white balancing algorithms. For both speed tests a C++ implementation of both algorithms was used on Windows 7 operating system on a computer with i5-2500K CPU and only one core was used. Color Sparrow was again used with parameters $N = 1, n = 225, r = 50, c = 50$ and with an averaging kernel of size $5 \times 5$. The algorithms were run several times on 100 images from the Shi's version of the Color Checker database. The average time for several runs of Gray-world algorithm was $2.9s$ for 100 images and the average time for CS was $3.03s$. These results show that even though CS is slower, it still performs almost as fast as Gray-world, but with more accuracy.

V. CONCLUSION AND FUTURE RESEARCH

Color Sparrow is a relatively fast and accurate method for white balancing. Even though in its core it is a modification of RSR, it calculates only the global illumination estimation and it also outmatches several other white balancing methods. It has the advantage of being unsupervised and performing well under lower sampling rates allowing a lower computation cost. This leads to conclusion that using RSR for global illumination estimation has a good potential of being a fast and accurate unsupervised color constancy method. However, as tests used in this paper were relatively simple, more exhaustive tests need to be performed in order to see if the accuracy can be improved even further. In future it would be good to test the proposed method on other color constancy databases and to perform experiments with other types of areas around particular pixels.

ACKNOWLEDGMENTS

The authors acknowledge the advice of Arjan Gijsenij on proper testing and comparison of color constancy algorithms.

This work has been supported by the IPA2007/HR/16IPO/001-040514 project "VISTA - Computer Vision Innovations for Safe Traffic."